\documentclass[runningheads]{llncs}

\usepackage{graphicx}
\usepackage{amsmath}
\usepackage{hyperref}
\usepackage{siunitx}
\usepackage[symbol]{footmisc}

\newcommand{\Embryoscope}{Embryoscope\textsuperscript{\textregistered{}}}
\newcommand{\micron}{\SI{}{\micro\meter}}
\newcommand{\boldparagraph}[1]{\textbf{#1:}}

\begin{document}

\title{Automated Measurements of Key Morphological Features of Human Embryos for IVF}
\titlerunning{Automated Measurement of Human Embryos}
\author{
    Brian D. Leahy\inst{*1, 2} \and Won-Dong Jang\inst{*1} \and Helen Y.
    Yang\inst{*3} \and Robbert Struyven\inst{1} \and Donglai Wei\inst{1}
    \and Zhe Sun\inst{1} \and Kylie R. Lee\inst{2} \and Charlotte
    Royston\inst{2} \and Liz Cam\inst{2} \and Yael Kalma\inst{4} \and
    Foad Azem\inst{4} \and Dalit Ben-Yosef\inst{4} \and Hanspeter
    Pfister\inst{1} \and Daniel Needleman\inst{1, 2}
}

%

\institute{
    School of Engineering and Applied Sciences,
    \and
    Department of Molecular and Cellular Biology,
    \and
    Harvard Graduate Program in Biophysics, \\
    Harvard University, Cambridge MA 02138, USA
    \and
    Tel Aviv Sourasky Medical Center, Tel Aviv, Israel
    \\ \email{bleahy@seas.harvard.edu}
}

\maketitle

\footnotetext[1]{These authors contributed equally to this work.}
\begin{abstract}

A major challenge in clinical In-Vitro Fertilization (IVF) is selecting the highest quality embryo to transfer to the patient in the hopes of achieving a pregnancy. Time-lapse microscopy provides clinicians with a wealth of information for selecting embryos. However, the resulting movies of embryos are currently analyzed manually, which is time consuming and subjective. Here, we automate feature extraction of time-lapse microscopy of human embryos with a machine-learning pipeline of five convolutional neural networks (CNNs). Our pipeline consists of (1) semantic segmentation of the regions of the embryo, (2) regression predictions of fragment severity, (3) classification of the developmental stage, and  object instance segmentation of (4) cells and (5) pronuclei. Our approach greatly speeds up the measurement of quantitative, biologically relevant features that may aid in embryo selection.

\end{abstract}

\section{Introduction}

One in six couples worldwide suffer from infertility~\cite{cui2010mother}. Many of those couples seek to conceive via In-Vitro Fertilization (IVF). In IVF, a patient is stimulated to produce multiple oocytes. The oocytes are retrieved, fertilized, and the resulting embryos are cultured \textit{in vitro}. Some of these are then transferred to the mother's uterus in the hopes of achieving a pregnancy; the remaining viable embryos are cryopreserved for future treatments. While transferring multiple embryos to the mother increases the potential for success, it also increases the potential for multiple pregnancies, which are strongly associated with increased maternal morbidity and offspring morbidity and mortality~\cite{norwitz2005maternal}. Thus, it is highly desirable to transfer only one embryo, to produce only one healthy child~\cite{practice2017guidance}. This requires clinicians to select the best embryos for transfer, which remains challenging~\cite{racowsky2011national}. 

The current standard of care is to select embryos primarily based on their morphology, by examining them under a microscope. In a typical embryo, after fertilization the two pronuclei, which contain the father's and mother's DNA, move together and migrate to the center of the embryo. The embryo undergoes a series of cell divisions, during the ``cleavage stage.'' Four days after fertilization, the embryo compacts and the cells firmly adhere to each other, at which time it is referred to as a compact ``morula.'' On the fifth day, the embryo forms a ``blastocyst,'' consisting of an outer layer of cells (the trophectoderm) enclosing a smaller mass (the inner-cell mass). On the sixth day, the blastocyst expands and hatches out of the zona pellucida (the thin eggshell that surrounds the embryo)~\cite{elder2020vitro}. Clinicians score embryos by manually measuring features such as cell number, cell shape, cell symmetry, the presence of cell fragments, and blastocyst appearance~\cite{elder2020vitro}, usually at discrete time points. Recently, many clinics have started to use time-lapse microscopy systems that continuously record movies of embryos without disturbing their culture conditions~\cite{rubio2014clinical,dolinko2017national,armstrong2019time}. However, these videos are typically analyzed manually, which is time-consuming and subjective.

Previous researchers have trained convolutional neural networks (CNNs) to directly predict embryo quality, using either single images or time-lapse videos \cite{petersen2016development,tran2019deep}. However, interpretability is vital for clinicians to make informed decisions on embryo selection, and an algorithm that directly predicts embryo quality from images is not interpretable. Worse, since external factors such as patient age~\cite{franasiak2014nature} and body-mass index~\cite{broughton2017obesity} also affect the success of an embryo transfer, an algorithm trained to predict embryo quality may instead learn a representation of confounding variables, which may change as IVF practices or demographics change. Some researchers have instead trained CNNs to extract a few identifiable features, such as blastocyst size~\cite{kheradmand2017inner}, blastocyst grade~\cite{kragh2019automatic,filho2012method,khosravi2019deep}, cell boundaries~\cite{rad2018hybrid}, or the number of cells when there are 4 or fewer~\cite{khan2016deep,lau2019embryo}. While extracting identifiable features obviates any problems with interpretability, these works leave out many key features that are believed to be important for embryo quality. Moreover, these methods are not fully automated, requiring the input images to be manually annotated as in the cleavage or blastocyst stage.

\begin{figure}[t]
\includegraphics[width=\textwidth]{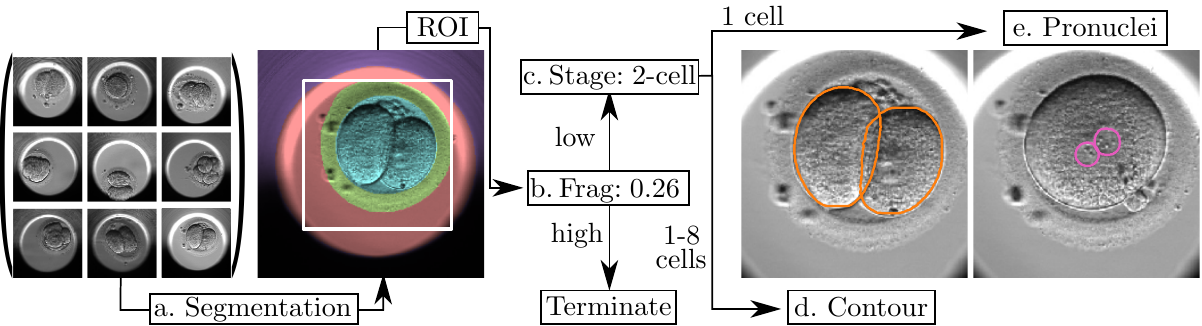}
\caption{
    Instead of performing one task, our unified pipeline extracts
    multiple features from embryos. We first segment the image to locate
    the embryo (panel~a), colored according to segmentation. The
    segmentation provides a region-of-interest (ROI, white box) for the
    other 4 networks, starting with embryo fragmentation (b); the
    image shown has a predicted fragmentation score of 0.26. If the
    embryo's fragmentation score is less than 1.5, we classify the
    developmental stage (c); this image is classified as a 2-cell
    embryo. We then detect cells in cleavage stage embryos (orange
    contours in d) and pronuclei in 1-cell embryos (magenta contours in
    e).
}
\label{fig:network}
\end{figure}

Here, we automate measurements of five key morphokinetic features of embryos in IVF by creating a unified pipeline of five CNNs. We work closely with clinicians to choose features relevant for clinical IVF: segmentation of the zona pellucida (Fig.~\ref{fig:network}a), grading the degree of fragmentation (Fig.~\ref{fig:network}b), classification of the developmental stage from 1-cell to blastocyst (Fig.~\ref{fig:network}c), object instance segmentation of cells in the cleavage stage (Fig.~\ref{fig:network}d), and object instance segmentation of pronuclei before the first cell division (Fig.~\ref{fig:network}e). With the exception of zona pellucida segmentation, all these features are used for embryo selection~\cite{alikani1999human,racowsky2011national,amir2019time,nickkho2019hydatidiform}; we segment the zona pellucida both to improve the other networks and because zona properties occasionally inform other IVF procedures~\cite{cohen1992implantation}. The five CNNs work together in a unified pipeline, combining results to improve performance over individual CNNs trained per task by several percent.

\section{Dataset}
We train the CNNs using data from the \Embryoscope{}, the most widely-used system for IVF with standardized, time-lapse microscopy~\cite{dolinko2017national}. \Embryoscope{} images are grayscale, and taken using Hoffman Modulation Contrast microscopy~\cite{hoffman1975modulation}, in which the intensity roughly corresponds to refractive index gradients. In our dataset, the \Embryoscope{} takes an image every 20 minutes at 7 focal planes, usually at 15~\micron{} increments. The recorded images provide views of the embryo with different amounts of defocus; they do not provide 3D information. The embryos are recorded for 3 -- 5 days, corresponding to 200 -- 350 images at each focal plane (\textit{i.e.}, 1400 -- 2450 images per embryo), although embryos are occasionally removed from the incubation system for clinical procedures. To train the CNNs, we curate a dataset with detailed, frame-by-frame labels for each task.

\section{Our Pipeline}
\label{sec:pipeline}
For each time-lapse video, we measure 5 morphokinetic features using 5 networks:

\boldparagraph{Zona Pellucida Segmentation} We first perform semantic segmentation to identify regions of the embryo, segmenting the image into four regions: pixels outside the well, inside the well, the zona pellucida, and the space inside the zona pellucida (the perivitelline space and embryo; Figure~\ref{fig:zona}, left). We segment the images by using a fully-convolutional network (FCN~\cite{long2015fully}; based on Resnet101~\cite{he2016deep}) to predict a per-pixel class probability for each pixel in the image. We train the FCN with images chosen from 203 embryos at 3,618 time points; we use neither a separate validation set nor early stopping for the zona segmentation.

The zona pellucida segmentation network in our pipeline takes the full 500$\times$500 pixel image as input. We use the segmentation result to crop the images to 328$\times$328, centered around the embryo, as input for the other networks.

\boldparagraph{Fragmentation Scoring} With the cropped image from the zona pellucida segmentation, we score the embryo's degree of fragmentation using a regression CNN (InceptionV3~\cite{szegedy2016rethinking}). The network takes a single-focus image as input and predicts a fragmentation score of 0 (0\% fragments), 1 ($<$10\%), 2 (10-20\%), or 3 ($\ge$20\%), following clinical practice. We train the network to minimize the $L^1$ loss on cleavage-stage images of 989 embryos at 16,315 times, each labeled with an integer score from 0--3; we use a validation set of 205 embryos labeled at 3,416 times for early stopping~\cite{goodfellow2016deep}. For each time point in the movie we analyze, we run the CNN on the three middle focal planes and take the average as the final score (Figure~\ref{fig:frag}, left).

Counting and identifying cells in fragmented embryos is difficult, inhibiting the labeling of train or test data for these embryos. Moreover, since high fragmentation is strongly correlated with low embryo viability~\cite{alikani1999human}, in standard clinical practice highly fragmented embryos are frequently discarded. Thus, we only train the rest of the networks on embryos with fragmentation less than 2.

\boldparagraph{Stage Classification} For low fragmentation embryos, we classify the embryo's developmental stage over time using a classification CNN (ResNeXt101~\cite{xie2017aggregated}). The classifier takes the three middle focal planes as input and predicts a 13-element vector of class probabilities, with 9 classes for cleavage-stage embryos (one each for 1--8 cells and one for $\ge 9$ cells) and one class each for morula (M), blastocyst (B), empty wells (E), and degenerate embryos (D; Figure~\ref{fig:stage}, left). To account for inaccuracies in the training data labels, we trained the classifier with a soft loss function modified from the standard cross-entropy loss
\begin{equation}
\log \left( p(\ell | m) \right) = \log \left( \sum_{t} p(\ell | t) p(t | m) \right) \quad ,
\label{eqn:softloss}
\end{equation}
where $t$ is the true stage of an image, $\ell$ the (possibly incorrect) label, and $m$ the model's prediction. We measured $p(\ell | t)$ by labeling 23,950 images in triplicate and using a majority vote to estimate the true label $t$ of each image. This soft-loss differs from the regularized loss in~\cite{szegedy2016rethinking} by differentially weighting classes; for instance, $p(\ell = \textrm{1-cell} | t = \textrm{1-cell}) = 0.996$ whereas $p(\ell = \textrm{6-cell} | t = \textrm{6-cell}) = 0.907$. Using the measured $p(\ell | t)$, we then trained the network with 341 embryos labeled at 111,107 times, along with a validation set of 73 embryos labeled at 23,381 times for early stopping~\cite{goodfellow2016deep}. Finally, we apply dynamic programming~\cite{bellman1966dynamic} to the predicted probabilities to find the most-likely non-decreasing trajectory, ignoring images labeled as empty or degenerate (Figure~\ref{fig:stage}, center).

\boldparagraph{Cell Object Instance Segmentation} For the images identified by the stage classifier as having 1--8 cells, we next perform object instance segmentation on each cell in the image. We train a network with the Mask R-CNN architecture~\cite{he2017mask} and a ResNet50 backbone~\cite{he2016deep}, using 102 embryos labeled at 16,284 times with 8 or fewer cells; we also use a validation set of 31 embryos labeled at 4,487 times for early stopping~\cite{goodfellow2016deep}. Our instance segmentation model takes as input a single-focus image cropped from the zona segmentation and resized to 500$\times$500. The segmentation model then predicts a bounding box, mask, and confidence score for each detected cell candidate (Figure~\ref{fig:contour}, left). Both the ground-truth labels and the predicted masks overlap significantly when the embryo has 2--8 cells (Figure~\ref{fig:contour}, center). We produce a final prediction by running our segmentation model on the three central focal planes; we merge candidates found across focal planes by using the one with the highest confidence score.

\boldparagraph{Pronucleus Object Instance Segmentation} Finally, in the images identified as 1-cell by the stage classifier, we detect the presence of pronuclei. To do so, we train another object instance segmentation network with the Mask R-CNN architecture~\cite{he2017mask} and a ResNet50 backbone~\cite{he2016deep}. We use a training set of 151 embryos labeled at 9,250 times during the 1-cell stage, with a validation set of 33 embryos labeled at 1,982 times for early stopping~\cite{goodfellow2016deep}. Pronuclei are only visible during a portion of the 1-cell stage; correspondingly, about 38\% of the training images contain 0, 6\% contain 1, and 54\% contain 2 pronuclei. The pronuclei detector takes as input a single image, cropped from the zona pellucida segmentation and resized to 500$\times$500, and it predicts a bounding box, mask, and confidence score for each detected candidate (Figure~\ref{fig:pronuclei}, left). We run the pronuclei detector on the three middle focal planes and merge candidates by using the one with the highest confidence score.

\section{Results}
We now evaluate our pipeline's performance, demonstrating the effect of each design choice in the models with ablation studies.

\begin{figure}
\includegraphics[width=\textwidth]{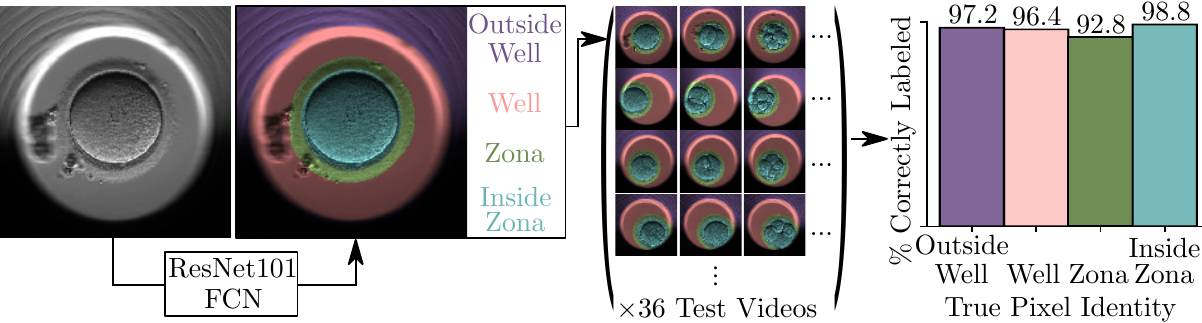}
\caption{
    The zona pellucida network (ResNet101 FCN) performs semantic segmentation
    on the input image, predicting four class probabilities for each
    pixel (colored as purple: outside well, pink: inside well, green:
    zona pellucida, cyan: inside zona). Middle: 12 representative
    segmentations of 3 embryos from the test set. Right: the per-pixel
    accuracies of the segmentation on each class in the test set.
}
\label{fig:zona}
\end{figure}

\boldparagraph{Zona Pellucida Segmentation} Our zona pellucida network nearly optimally segments the test set images, taken from 36 embryos at 576 times. The FCN correctly labels image pixels 96.7\% of the time, with per-class accuracies between 93-99\% (Figure~\ref{fig:zona}, right). The misclassified pixels arise mostly at region boundaries, roughly corresponding to the few-pixel human labeling inprecision at region boundaries.

\begin{figure}
\includegraphics[width=\textwidth]{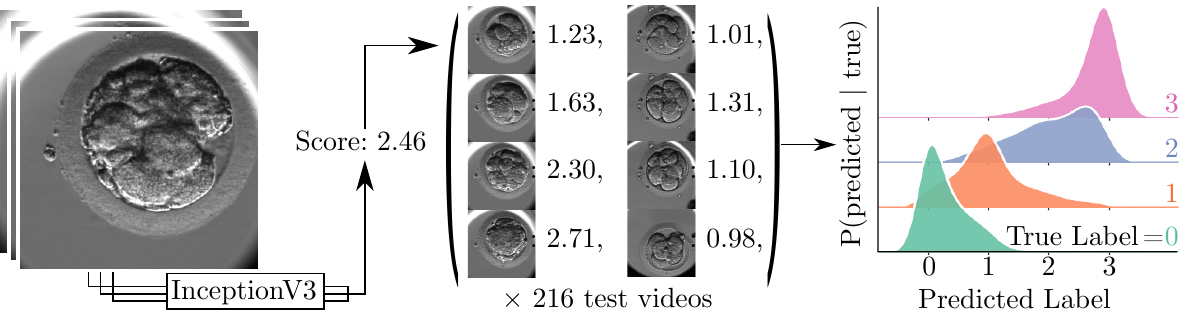}
\caption{
    Left: The fragmentation network (InceptionV3 architecture) scores
    embryos with a real number from 0 -- 3; the image at left is scored
    as a fragmentation of 2.46. Center: 8 representative fragmentation
    scores on the test set, shown as image: score pairs.  Right: The
    distribution of the network's prediction given the ground-truth
    label on the test set. The green distribution corresponds to images
    with a ground-truth label of 0; orange those labeled as 1; blue, 2;
    pink, 3.
}
\label{fig:frag}
\end{figure}

\boldparagraph{Fragmentation Scoring} The network predicts a score with a mean-absolute deviation of 0.45 from the test labels on the fragmentation test set of 216 embryos labeled at 3,652 times (Figure~\ref{fig:frag}, right). When distinguishing between low- ($<1.5$) and high- ($\ge 1.5$) fragmentation, the network and the test labels agree 88.9\% of the time. Our network outperforms a baseline InceptionV3 by 1.9\%; focus averaging and cropping to a region-of-interest each provide a 1--1.5\% boost to the accuracy (Table~\ref{table:ablation}).

We suspect that much of the fragmentation network's error comes from imprecise human labeling of the train and test sets, due to difficulties in distinguishing fragments from small cells and due to grouping the continuous fragmentation score into discrete bins. To evaluate the human labeling accuracy, two annotators label the fragmentation test set in duplicate and compare their results. The two annotators have a mean-absolute deviation of 0.37 and are 88.9\% consistent in distinguishing low- from high- fragmentation embryos. Thus, the fragmentation CNN performs nearly optimally in light of the labeling inaccuracies.

\begin{figure}
\includegraphics[width=\textwidth]{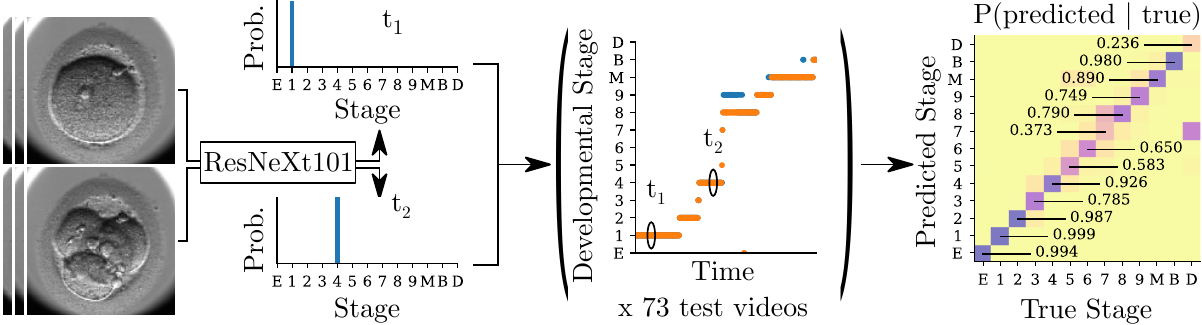}
\caption{
    Left: The stage classification CNN (ResNeXt101) predicts a per-class
    probability for each image; the two bar plots show the predicted
    probabilities for the two images. Center: We use dynamic
    programming to find the most-likely non-decreasing trajectory
    (orange); the circled times $t_1$ and $t_2$ correspond to the
    predictions at left. Right: The distribution of predictions given
    the true labels, measured on the test set.
}
\label{fig:stage}
\end{figure}

\boldparagraph{Stage Classification} The stage classifier predicts the developmental stage with a 87.9\% accuracy on the test set, consisting of 73 embryos labeled at 23,850 times (Figure~\ref{fig:stage}, right). The network's accuracy is high but lower than the human labeling accuracy on the test set (94.6\%). The network outperforms a baseline ResNeXt101 by 6.7\%; both the soft-loss and the dynamic programming each improve the predictions by 2\% (Table~\ref{table:ablation}). The stage classifier struggles when there are between 5 and 8 cells (66.9\% accuracy for these classes). In contrast, the stage classifier does exceedingly well on images with 1-cell (99.9\%), 2-cells (98.7\%), empty wells (99.4\%), or blastocysts (98.0\%; Figure~\ref{fig:stage}, right). Despite measuring significantly more developmental stages, our stage classifier outperforms previous cell counting networks developed for human embryos~\cite{khan2016deep,lau2019embryo}.

\begin{figure}
\includegraphics[width=\textwidth]{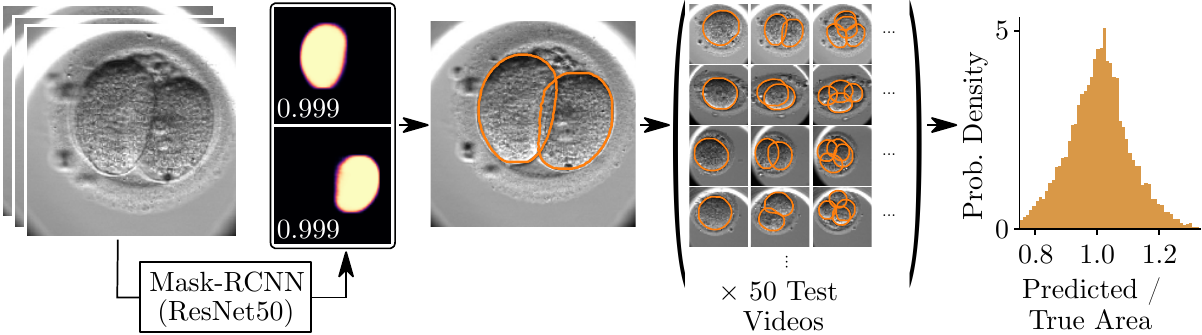}
\caption{
    The cell detection network (Mask-RCNN, ResNet50 backbone) takes an
    image (left) and proposes candidates as a combined object mask and
    confidence score from 0--1 (second from left). Center: The boundaries of
    the object mask represented as the cell's contours (orange, center).
    Second from right: 12 cell instance segmentations for 4 embryos from
    the test set (shown as orange contours overlaid on the original
    image). Right: Histogram of the ratio of predicted to
    true areas for correctly identified cells in the test set.
}
\label{fig:contour}
\end{figure}

\begin{table}
\centering
\caption{
    Effect of design choices for the 4 more difficult tasks, illustrated
    by removing one modification to the network at a time. Test-set
    scores are in percent correctly classified (stage, fragmentation)
    and mean-average precision (blastomere, pronuclei). The best scores
    are boldfaced. The last row shows the test set scores using all the
    training data but no input from other networks and no
    modifications to the network.
}
\label{table:ablation}
\begin{tabular}{c|c|c|c|c}
\hline
 & Fragmentation & \, Stage \, & Blastomere & Pronuclei \\
Setting & (\%) & (\%) & (mAP) & (mAP)\\
\hline
\hline
Full Setting            & \textbf{88.9} & \textbf{87.8} & 0.737     & \textbf{0.680} \\ \hline
Single Focus            &  87.8         & 84.8          & \textbf{0.739}    & 0.668 \\ \hline
No ROI from Zona        &  87.4         & 84.9          & 0.733             & 0.666 \\ \hline
Using 50\% Training Data & 87.7        & 85.3          & 0.718             & 0.656 \\ \hline
No Soft Loss            &  --           & 85.3          & --                & --    \\ \hline
No Dynamic Programming  &  --           & 86.0          & --                & --    \\ \hline \hline
Single-Task Baselines      & 87.0  & 81.1  & 0.737  & 0.647 \\
                        & \cite{szegedy2016rethinking} & \cite{xie2017aggregated} & \cite{he2017mask} & \cite{he2017mask} \\ \hline
\end{tabular}
\end{table}

\boldparagraph{Cell Object Instance Segmentation} We measure the accuracy of the cell instance segmentation network using mean-average precision (mAP)~\cite{lin2014microsoft}, a standard metric for object instance segmentation tasks. Our network predicts cell masks with a mAP of 0.737 on the test set, consisting of 31 embryos labeled at 4,953 times. The model identifies cells with a precision of 82.8\% and a recall of 88.4\%, similar to results from other work on fewer images~\cite{rad2018hybrid}. For correctly-identified candidates, the predicted cell area is within 17\% of the true cell area 90\% of the time (Figure~\ref{fig:contour}, right); much of this error arises when cells strongly overlap late in the cleavage stage. Cropping to a region-of-interest provides a marginal improvement to the network's accuracy (Table~\ref{table:ablation}).

\begin{figure}
\includegraphics[width=\textwidth]{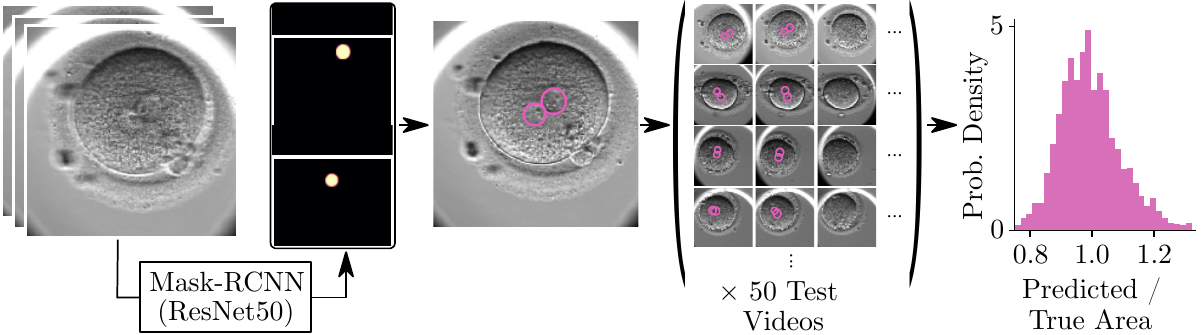}
\caption{
    The pronuclei detection network (Mask-RCNN, ResNet50 backbone) takes an
    image (left) and proposes candidates as a combined object mask and
    confidence score from 0--1 (second from left). Center: The boundaries of
    the object mask represented as the pronuclei contours (magenta, center).
    Second from right: 12 pronuclei instance segmentations for 4 embryos from
    the test set (shown as magenta contours overlaid on the original
    image); the rightmost images illustrate true negatives after the
    pronuclei have faded. Right: Histogram of the ratio of predicted to
    true areas for correctly identified pronuclei in the test set.
}
\label{fig:pronuclei}
\end{figure}

\boldparagraph{Pronucleus Object Instance Segmentation} The pronuclei segmentation network predicts masks with a mAP of 0.680 on the test set of 33 embryos labeled at 2,090 times. The network identifies pronuclei with a precision of 81.4\% and a recall of 88.2\%. Much of the false positive detections are from vacuoles inside the 1-cell embryo, which look similar to pronuclei. For correctly-identified candidates, the predicted pronuclei area is within 16\% of the true pronuclei area 90\% of the time (Figure~\ref{fig:pronuclei}, right). The pronuclei network's mAP outperforms that of a baseline Mask-RCNN by 0.03; averaging across focal planes and cropping to a region-of-interest each improves the mAP by 0.01 (Table~\ref{table:ablation}).

\section{Conclusions}

Our unified pipeline greatly speeds up the measurement of embryos: running all five networks on a 300-image, five-day movie takes 6 minutes on a GTX Titan X. In the future, we plan to make this pipeline even faster by combining all five networks with multi-task learning~\cite{dai2016instance}. Since we measure many of the key morphological features used in clinical IVF, our unified pipeline has the potential to reduce the time to grade embryos without sacrificing interpretability. Equally as important, the automatic, high-quality data produced by our pipeline will enable better retrospective chart studies for IVF, improving IVF by informing better clinical practice.

\section*{Acknowledgements}
We acknowledge M. Venturas and P. Maeder-York for help validating labels and approaches. This work was funded in part by NIH grant 5U54CA225088 and NSF Grant NCS-FO 1835231, by the NSF-Simons Center for Mathematical and Statistical Analysis of Biology at Harvard (award number 1764269), and by the Harvard Quantitative Biology Initiative. DJN and DBY also acknowledge generous support from the Perelson family, which made this work possible.

\bibliographystyle{splncs04}
\bibliography{miccai_bibliography}

\end{document}